\documentclass[letterpaper, 10 pt, conference]{ieeeconf}  

\IEEEoverridecommandlockouts                              

\usepackage{graphics} 
\usepackage{epsfig} 
\usepackage{mathptmx} 
\usepackage{times} 
\usepackage{amsmath} 
\usepackage{amssymb}  

\usepackage{bbding}

\usepackage[utf8]{inputenc} 
\usepackage[T1]{fontenc}    
\usepackage{hyperref}       
\usepackage{url}            
\usepackage{booktabs}       
\usepackage{amsfonts}       
\usepackage{nicefrac}       
\usepackage{microtype}      
\usepackage{color}
\usepackage[table]{xcolor}
\usepackage{cleveref}
\usepackage{graphicx}
\usepackage{float}
\usepackage{multirow}
\usepackage{diagbox}

\usepackage{algorithm}
\usepackage{algpseudocode}
\usepackage{arydshln}
\usepackage{colortbl}

\usepackage{multicol}
\usepackage{multirow}
\usepackage{listings}
\definecolor{dkgreen}{rgb}{0,0.6,0}
\definecolor{gray}{rgb}{0.5,0.5,0.5}
\definecolor{mauve}{rgb}{0.58,0,0.82}
\usepackage{tikz}
\newcommand*\emptyCircle[1][0.75ex]{\tikz\draw (0,0) circle (#1);} 
\newcommand*\fullCircle[1][0.75ex]{\tikz\fill (0,0) circle (#1);} 
\newcommand*\halfCircle[1][0.75ex]{%
	\begin{tikzpicture}
	\draw[fill] (0,0)-- (90:#1) arc (90:270:#1) -- cycle ;
	\draw (0,0) circle (#1);
	\end{tikzpicture}}

\title{\LARGE \bf
Unsupervised Spike Depth Estimation via Cross-modality Cross-domain Knowledge Transfer
}

\author{Jiaming Liu$^{1*}$, Qizhe Zhang$^{1*}$, Xiaoqi Li$^{1}$, Jianing Li$^{1}$, Guanqun Wang$^{1}$~\textsuperscript{\Envelope}, \\ Ming Lu$^{1}$, Tiejun Huang$^{1}$, Shanghang Zhang$^{1}$~\textsuperscript{\Envelope}
\thanks{$^1$ Jiaming Liu, Qizhe Zhang, Xiaoqi Li, Jianing Li, Guanqun Wang, Ming Lu, Tiejun Huang and Shanghang Zhang are with National Key Laboratory for Multimedia Information Processing, School of CS, Peking University. *: Equal Contribution: jiamingliu@stu.pku.edu.cn; theia4869@gmail.com.  \textsuperscript{\Envelope}:
Corresponding Author: shanghang@pku.edu.cn; wgq@pku.edu.cn}
}

\begin{document}
\maketitle
\begin{abstract}

Neuromorphic spike data, an upcoming modality with high temporal resolution, has shown promising potential in autonomous driving by mitigating the challenges posed by high-velocity motion blur. However, training the spike depth estimation network holds significant challenges in two aspects: sparse spatial information for pixel-wise tasks and difficulties in achieving paired depth labels for temporally intensive spike streams. Therefore, we introduce open-source RGB data to support spike depth estimation, leveraging its annotations and spatial information. The inherent differences in modalities and data distribution make it challenging to directly apply transfer learning from open-source RGB to target spike data. To this end, we propose a cross-modality cross-domain (BiCross) framework to realize unsupervised spike depth estimation by introducing simulated mediate source spike data. Specifically, we design a Coarse-to-Fine Knowledge Distillation (CFKD) approach to facilitate comprehensive cross-modality knowledge transfer while preserving the unique strengths of both modalities, utilizing a spike-oriented uncertainty scheme. Then, we propose a Self-Correcting Teacher-Student (SCTS) mechanism to screen out reliable pixel-wise pseudo labels and ease the domain shift of the student model, which avoids error accumulation in target spike data. To verify the effectiveness of BiCross, we conduct extensive experiments on four scenarios, including Synthetic to Real, Extreme Weather, Scene Changing, and Real Spike. Our method achieves state-of-the-art (SOTA) performances, compared with RGB-oriented unsupervised depth estimation methods. Code and dataset: \href{https://github.com/Theia-4869/BiCross} {https://github.com/Theia-4869/BiCross}.

\end{abstract}

\section{Introduction}

\begin{figure}[t]
\includegraphics[width=0.48\textwidth]{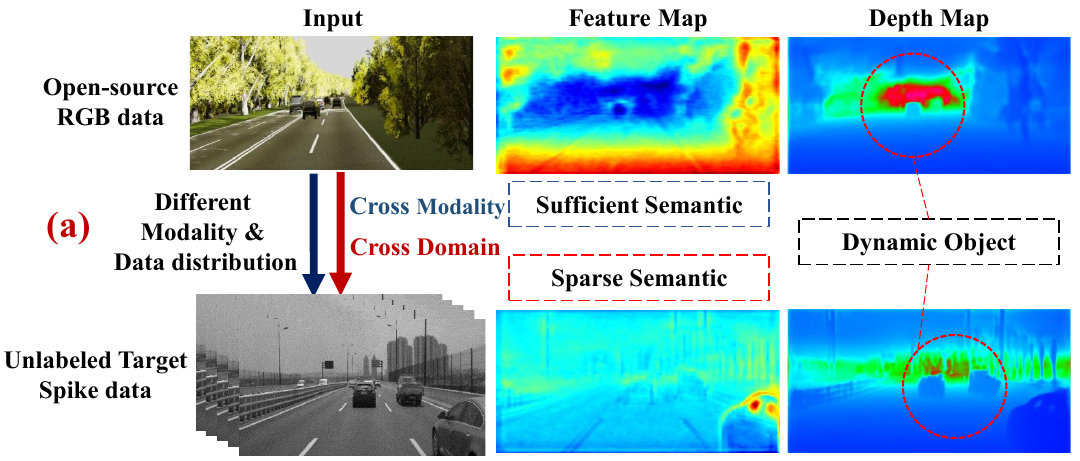}
\centering
\vspace{-0.65cm}
\caption{demonstrates the process of BiCross and the distinct properties of RGB and spike modality. While RGB data contains sufficient semantic information, spike data is limited in its feature space due to sparse spatial information. However, spike data shows strength in depth estimation on dynamic objects, owning to its dense temporal resolution.}
\label{fig:intro}
\vspace{-0.4cm}
\end{figure}

The neuromorphic spike camera generates data streams with high temporal resolution in a bio-inspired way~\cite{dong2019efficient,zhu2019retina}, which has shown promising potential in real-world applications, such as autonomous driving~\cite{manhardt2019roi,wu20196d, chi2023bev, li2023bev, yang2023lidar} and robotic manipulation~\cite{tremblay2018deep, li2023manipllm, li2023imagemanip}.
The spike camera has an inherent advantage over the RGB camera, which has severe performance degradation in the high-velocity scenario because of motion blur~\cite{hu2021optical}.
Therefore, as shown in Fig.~\ref{fig:intro}, we attempt to perform depth estimation on spike data, which shows strength on dynamic objects~\cite{hu2014joint}.

However, conducting depth estimation on spike data is difficult due to the intrinsic properties of spike data: (1) \textit{Sparse spatial information.} Since the spike camera adopts the firing mechanism to capture pixel-wise luminance intensity, it may miss some interactions, leading to sparse spatial information~\cite{dong2021spike, li2022retinomorphic, zhang2022spike}.
(2) \textit{Dense temporal streams.} Spike data is captured with high-frequency frames up to  40000 Hz, resulting in extraordinarily dense temporal frames~\cite{dong2021spike,hu2021optical}. Though the property can better perceive objects of high-velocity motion blur, the annotation process becomes extremely laborious. 
In this paper, we make the first attempt to organize a study of unsupervised spike depth estimation. Though its RGB-oriented counterpart has been studied, directly applying these methods on spike modality will lead to performance degradation since they are driven by geometric or content consistency~\cite{godard2017unsupervised,lopez2020desc} and are not feasible for sparse spike streams. Therefore, we introduce open-source RGB data to assist spike depth estimation by exploiting its annotation and absorbing sufficient spatial information.

To reconcile such, we propose a cross-modality cross-domain (BiCross) framework for unsupervised spike depth estimation which introduces a simulated mediate source spike data and breaks the difficulties of transfer learning into steps.
In the cross-modality phase, we propose a  Coarse-to-Fine Knowledge Distillation (CFKD) to transfer sufficient semantic knowledge from RGB to source spike data. It adopts a pixel-wise uncertainty filter to screen out the distilled knowledge which should be reliable in RGB features and demanded in sparse spike features, thus reserving the unique strength of both modalities and complementing the spatial information of spike data.
In the cross-domain phase, we further introduce a Self-Correcting Teacher-Student (SCTS) mechanism to exploit transferred cross-modality knowledge and better address the domain shift between simulated source spike and target spike data. Specifically, due to the sparse property of spike data and the domain shift, it will usually lead to unreliable predictions. We thus screen out more reliable pixel-wise pseudo labels to guide cross-domain learning and avoid error accumulation.

We conduct extensive experiments to demonstrate that our method achieves competitive performance on the unsupervised spike depth estimation task. We design four BiCross scenarios, which are \textbf{Synthetic to Real} (Virtual KITTI RGB~\cite{gaidon2016virtual} to KITTI spike), \textbf{Extreme Weather} (clear RGB~\cite{yang2019drivingstereo}  to foggy spike), \textbf{Scene Changing} (KITTI~\cite{geiger2012we} RGB to Driving Stereo spike), and \textbf{Real Spike} (NYUv2 RGB~\cite{silberman2012indoor} to Respike indoor spike~\cite{wang2022respike}) scenario respectively. 
The main contributions are summarized as follows:

\textbf{1)} We propose a cross-modality cross-domain (BiCross) framework for unsupervised spike depth estimation, and make the first attempt to exploit the open-source RGB datasets to assist spike modality tasks by leveraging RGB annotation and absorbing sufficient spatial information. 

\textbf{2)} In the BiCross framework, we introduce a Coarse-to-Fine Knowledge Distillation (CFKD) and a Self-Correcting Teacher-Student (SCTS) mechanism to realize transfer learning from open-source RGB to target spike datasets. CFKD reserves the unique strength of both modalities and complements the spatial information of sparse spike data. SCTS further addresses the domain shift between the simulated source and the target spike data in an unsupervised manner.

\textbf{3)} We achieve SOTA performances on four challenging BiCross scenarios, compared with RGB-oriented unsupervised depth estimation methods. We provide four large-scale spike datasets for continuous research in the spike community.

\section{Related Work}
\textbf{Monocular depth estimation} \quad 
Depth estimation is an important task of machine scene understanding. Deep learning has become a prevailing solution to supervised depth estimation for both outdoor~\cite{eigen2014depth, geiger2012we, yang2019drivingstereo} and indoor~\cite{Silberman:ECCV12, scharstein2014high} scenes.
These methods usually consist of a general encoder to extract global context information and a decoder to recover depth information~\cite{Xu_2018_CVPR, Ramamonjisoa_2020_CVPR, lee2019monocular, Ramamonjisoa_2019_ICCV, fu2018deep}. 
However, the supervised methods need a mass of annotation in pixel-level thus limiting their scalability and practicability. As for spike stream, it is nearly impossible to train neural models in a supervised manner since spike stream is too temporally intensive to obtain paired depth labels.
In contrast, unsupervised depth estimation methods do not require ground-truth depth to train the models~\cite{garg2016unsupervised,zhou2017unsupervised,godard2017unsupervised,godard2019digging,pilzer2019refine,chen2021revealing}. 
Existing unsupervised RGB methods are driven by photometric consistency, which will cause severe performance degradation when directly applied to spike streams.

\textbf{Adaptive depth estimation} \quad 
Existing cross-modality depth estimation methods usually take advantage of aligned data from different modalities~\cite{piao2021learning,verdie2022cromo}.
~\cite{wang2021evdistill} proposes to use aligned DVS event data and APS images to perform cross-modality knowledge distillation for depth estimation with unaligned event data. 
Cross-domain depth estimation usually aligns the source and target domains from input level or feature level~\cite{kundu2018adadepth,zheng2018t2net,zhao2019geometry}.
T2Net \cite{zheng2018t2net} and \cite{zhao2019geometry} developed an end-to-end and geometry-aware symmetric adaptation framework respectively, which optimize the translation and the depth estimation network.

\begin{figure*}[t]
\includegraphics[width=0.95\textwidth]{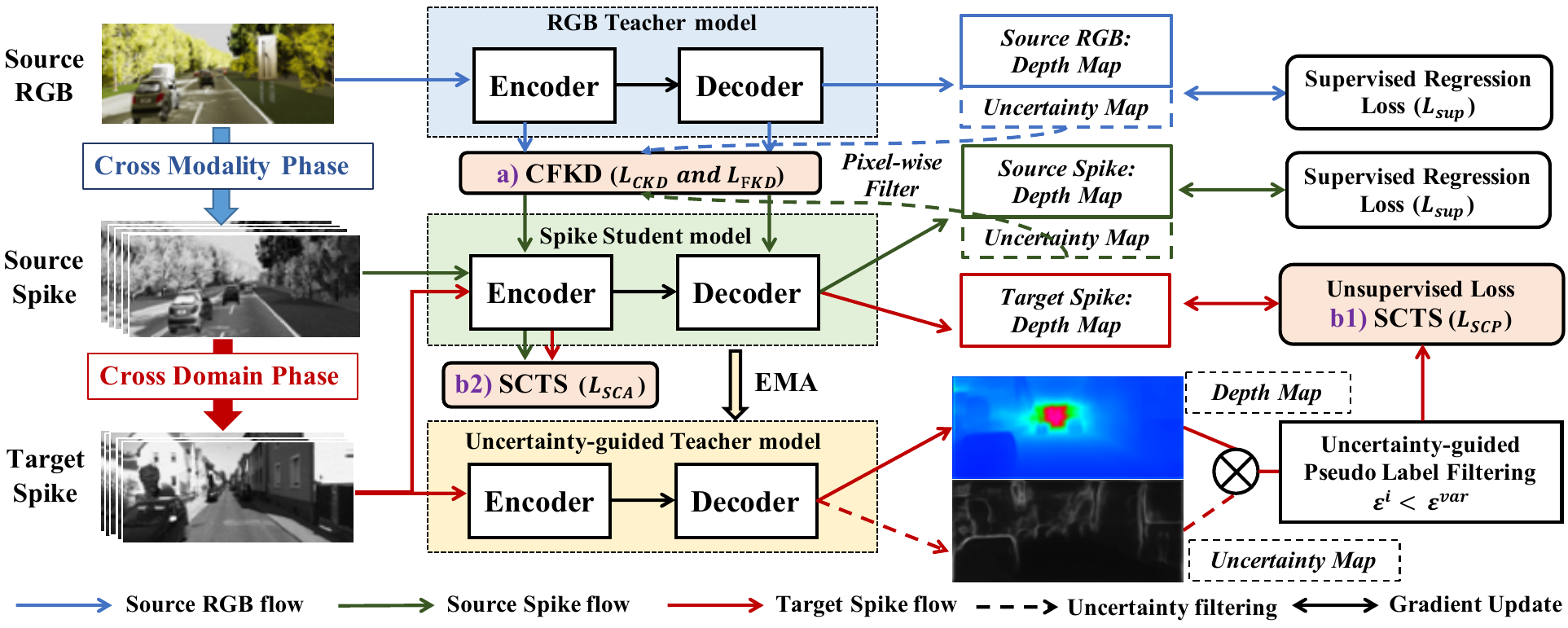}
\centering
\vspace{-0.39cm}
\caption{The BiCross framework composes the cross-modality and cross-domain phases. The first two rows demonstrate cross-modality learning, we propose CFKD with the spike-oriented uncertainty filter (part a) to transfer sufficient knowledge from \textit{RGB teacher model} to \textit{spike student model} under the source domain. The last two rows show the cross-domain learning,  we introduce a Self-Correcting Teacher-Student (SCTS) scheme in which the teacher model utilizes the pixel-wise uncertainty method to screen out reliable depth estimation (part b1) and the student model adopts global-level alignment (part b2) to correct domain shift. The uncertainty estimation approach is conducted at the model output layer, following the two phases. 
CFKD and SCTS jointly contribute to achieving unsupervised spike depth estimation.}
\label{fig:method}
\vspace{-0.5cm}
\end{figure*}

\textbf{Spike camera and its application} \quad 
Spike camera is a kind of bio-inspired sensor~\cite{dong2021spike} with high temporal resolution. Based on obtained spike frames from spike camera, existing works concentrate on spike-to-image reconstruction~\cite{zhu2020retina,9181055,zheng2021high,zhu2021neuspike, zhao2021super, zhao2021spk2imgnet, zhu2022ultra}, which takes advantage of high temporal resolution of spike streams and generates high SNR as well as high frequency reconstructed images. Meanwhile, ~\cite{hu2021scflow} proposes SCFlow to predict high-speed optical flow from spike streams. ~\cite{li2022retinomorphic} proposes a retinomorphic object detection method to fuse the DVS modality and spike modality via a dynamic interaction mechanism.
Compared with DVS~\cite{delbruck2010activity,lichtsteiner2008128}, spike camera adopts the firing operation to capture the pixel-wise luminance intensity instead of pixel-wise luminance change. Therefore, although both cameras are capable of reserving high temporal resolution, spike camera has more advantages for high-speed depth estimation in regions with weak and boundary textures~\cite{hu2021optical}.
In this paper, we make the first attempt to explore the unsupervised depth estimation for spike data and leverage the individual property of RGB and spike modalities.

\section{Method}

\subsection{Preliminary and Overall}
\label{sec:3.1}

\textbf{Neuromorphic spike data} \quad
Spike camera utilizes photo receptions to capture natural lights which are converted to voltage under integration of time series $t$. Once the voltage at a certain sensing unit reaches a threshold $\Theta$, a one-bit spike is fired and the voltage is reset to zero~\cite{9181055}.
\begin{equation}
\scriptstyle
   S(i, j, t) = \left\{
    \begin{array}{l}
            1 , \quad \int_{{t_0}_{i,j}^{pre}}^{t} I(i, j) \, dt \ge \Theta\\  
            0 , \quad \int_{{t_0}_{i,j}^{pre}}^{t} I(i, j) \, dt < \Theta
        \end{array}
\right.
\label{eq1}
\end{equation}
The Eq. \ref{eq1} reveals the basic working pipeline of the spike camera, where $I(i,j)$ represents the luminance of pixel $(i,j)$ and ${t_0}^{pre}_{i,j}$ represents the time that fires the last spike at pixel$(i,j)$. The spike fires when the accumulation of luminance reaches the threshold, capturing high frequency frames up to 40000 Hz~\cite{zhu2020retina, Zhao_2021_CVPR}. Following previous works~\cite{hu2021optical, zhu2021neuspike}, we simulate the spike frames from open source datasets under the work flow of spike camera. Then, we split the spike sequence into streams $I^{spike} = H \times W \times T$, $H$ and $W$ represent the height and width of spike frames, under the temporal resolution $T$~\cite{wang2022learning,zhang2022spike}.

\begin{figure*}[t]
\includegraphics[width=1.0\textwidth]{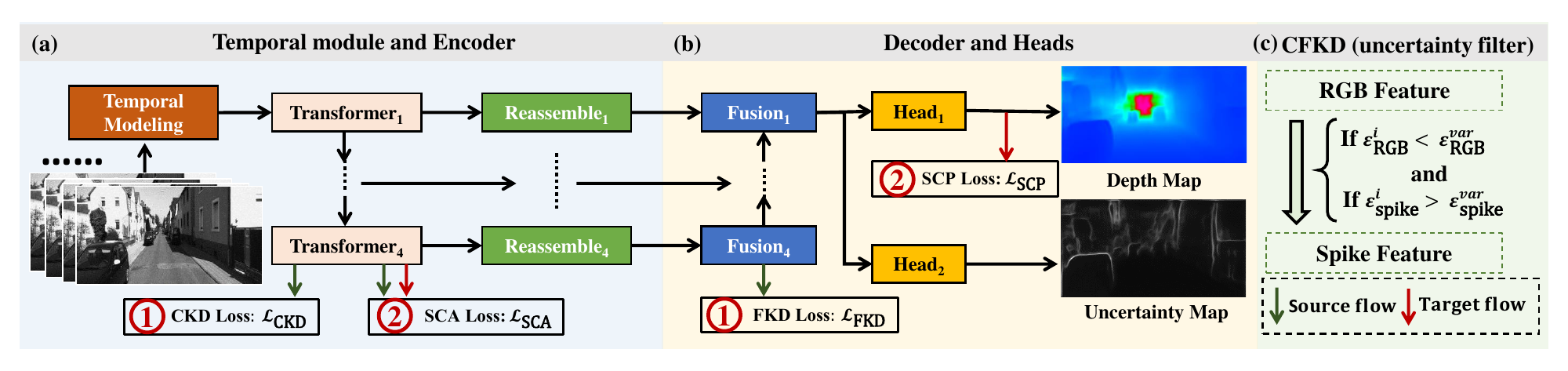}
\centering
\vspace{-0.9cm}
\caption{The spike data is sent into the spike encoder as shown in part (a). Part (b) shows the decoder and prediction heads of the network. In part (c), we show the mechanism of spike-oriented uncertainty filter in CFKD, it selects the distilled knowledge which should be reliable in the RGB feature and demanded in the sparse spike feature. \textcolor{red}{Red circle.1} represents the objectives of cross-modality learning, including coarse (CKD) and fine-level (FKD) knowledge distillation. \textcolor{red}{Red circle.2} contains self-correcting pseudo-label (SCP) and self-correcting alignment (SCA) in the cross domain phase. }
\label{fig:IGKD}
\vspace{-0.5cm}
\end{figure*}

\textbf{Neuromorphic spike network} \quad
In this paper, both RGB and spike modality embed features with the help of DPT~\cite{ranftl2021vision}, where the input data are spike  $I^{spike} \in \mathbb{R}^{1280 \times H \times W}$ and RGB $I^{rgb} \in \mathbb{R}^{3 \times H \times W}$. As shown in Fig.~\ref{fig:IGKD} (a), the network encoder is divided into two blocks, the first one is a temporal modeling module (temporal attention and Resblocks) to adaptively aggregate the dense temporal information of special spike data, which is utilized to reduce computational cost when the spike temporal resolution is too high. 
The second one is ViT-Hybrid, which uses a ResNet-50~\cite{he2016deep} to encode the image and spike embedding followed by 12 transformer layers. In Fig.~\ref{fig:IGKD} (b), the decoder consists of the reassemble operation, along with the fusion module.
Fig.~\ref{fig:IGKD} (c) contains two prediction heads, where $Head_1$ predicts depth and $Head_2$ estimates corresponding uncertainty map for each input.

\textbf{BiCross framework}
We propose a cross-modality cross-domain (BiCross) framework (shown in Fig .\ref{fig:method}) by introducing an intermediate source spike domain. The key insight is to break the complicated transferring process step by step and facilitate unsupervised spike depth estimation.

\subsection{Coarse-to-Fine Knowledge Distillation}
\label{sec:cfkd}
Since depth estimation is a pixel-wise task while spike data lack intact spatial information, we intend to transfer sufficient semantic knowledge from RGB to spike modality. As shown in the top of Fig.~\ref{fig:method}, we propose a Coarse-to-Fine Knowledge Distillation (CFKD) with a pixel-wise uncertainty filter to transfer the comprehensive semantic knowledge from RGB model $\mathcal{T}_{RGB}$ to spike model $\mathcal{S}_{src}$. The uncertainty filter is specially designed to select information that is reliable in RGB modality and demanded in sparse spike modality. The goal is to reserve the unique strength of both modalities and provide more available spatial information for the following cross-domain transferring.

\textbf{Spike-oriented Uncertainty filter} \quad 
Since the spike data capture scene information by pixel luminance intensity, the spike streams present sparse spatial information at the pixel level.
Therefore we propose a pixel-wise uncertainty filter, which is utilized in both cross-modality and cross-domain phases. 
Specifically, we generate soft labels for uncertainty measurement as follows:
\begin{equation}
\scriptstyle
     \mathcal{E}_{soft} = \frac{|\mathbf{D}_{pred}-\mathbf{D}_{gt}|}{\mathbf{D}_{gt}}
\end{equation}
where $\mathbf{D}_{pred}$ represents the output of depth estimation head and $\mathbf{D}_{gt}$ represents the depth ground-truth (in cross-modality phase) or pseudo-label (in cross-domain phase). 
Since we observe uncertainty degree is positively correlated with the disparity of prediction and ground truth, soft labels can serve as the supervision for uncertainty estimation. 
The uncertainty map $\mathbf{U}$ and $\mathcal{E}_{soft}$ are further penalized by uncertainty L1 loss ($\mathcal{L}_{\text{unc}}$) in the two-phase learning.
As shown in Fig.~\ref{fig:IGKD} (c), we distill the knowledge of a pixel if its RGB modality prediction uncertainty is below the threshold $\mathcal{E}_{\text{RGB}}^{var}$ and spike uncertainty prediction is above the threshold $\mathcal{E}_{\text{spike}}^{var}$. 
It thus ensures the transferred knowledge is reliable and demanded.

\textbf{Knowledge distillation} \quad 
We then aim to distill the selected knowledge to student model with both coarse (CKD) and fine (FKD) knowledge distillation strategies.
In the CKD, we first obtain the high-dimensional feature $\mathbf{F}_{enc}\in\mathbb{R}^{C_e \times H_e \times W_e}$ from the encoder of the model and utilize an average pooling followed by a MLP to aggregate it into a global-level representation vector $\mathbf{f}_{g}=\text{MLP}(\text{Pool}_{avg}(\mathbf{F}_{enc}))\in\mathbb{R}^{C_g}$. Then, the KL Divergence loss is used to optimize coarse-grained distillation.
Since depth estimation is a pixel-wise regression task, we adopt FKD to further align the pixel-level features of the two modalities. We get the feature map $\mathbf{F}_{dec}\in\mathbb{R}^{C_d \times H_d \times W_d}$ from the decoder of the model and then directly adopt the MSE loss to achieve feature distillation. 
As shown in Fig.~\ref{fig:IGKD}, CKD and FKD are applied in the encoder and decoder respectively while their loss functions are as follows:
\begin{equation}
\scriptstyle
\mathcal{L}_{\text{CFKD}} = \mathcal{L}_{\text{CKD}}+\mathcal{L}_{\text{FKD}} = \mathcal{D}_{\text{KL}
}(\hat{\mathbf{f}}_{g}^{\mathcal{T}}\|\hat{\mathbf{f}}_{g}^{\mathcal{S}})+\frac{1}{H_d \times W_d}\|\mathbf{F}_{dec}^{\mathcal{T}}-\mathbf{F}_{dec}^{\mathcal{S}}\|^{2}
\label{eq:CFKD}
\end{equation}
where $\mathcal{T}$ and $\mathcal{S}$ denote the feature from teacher and student model, $\hat{\mathbf{f}}_{g}^{\mathcal{T}}$ and $\hat{\mathbf{f}}_{g}^{\mathcal{S}}$ stand for normalized global-level representation vectors. $\mathbf{F}_{enc}^{\mathcal{T}}, \mathbf{F}_{enc}^{\mathcal{S}}, \mathbf{F}_{dec}^{\mathcal{T}}$ and $\mathbf{F}_{dec}^{\mathcal{S}}$ in this section are selected by the spike-oriented uncertainty map which is reshaped to the same resolution as the features.

\subsection{Self-Correcting Teacher-Student Mechanism}
After obtaining a comprehensive source spike representation, we then aim to pull close the data distribution of the source spike and target spike domain and realize unsupervised spike depth estimation. Though the widely-used teacher-student mechanism can ease domain shift through generating target domain pseudo label~\cite{yu2022cross, pham2021meta}, it will result in vast unreliable pixel-wise pseudo labels due to the sparse property of the spike modality. This motivates us to propose a Self-Correcting Teacher-Student (SCTS) framework in which the teacher model utilizes a spike-oriented uncertainty scheme to screen out reliable depth estimation and the student model adopts global-level alignment to correct domain shift (bottom of Fig.~\ref{fig:method}). In this way, we avoid error accumulation in the framework and better address domain shifts in spike modality.

\textbf{Self-correcting pseudo-label} \quad 
The initial weights of teacher $\mathcal{T}_{mean}$ and student models $\mathcal{S}_{tgt}$ are loaded from the source pre-trained model.
The target student model is updated with back-propagation, and the teacher model is updated by student's weights with exponential moving average (EMA) \cite{tarvainen2017mean}. The weights of the teacher model at time step $t$ can be expressed as:
\begin{equation}
\scriptstyle
\label{eq:2}
     \mathcal{T}_{mean}^{t} = \alpha \mathcal{T}_{mean}^{t-1} + (1-\alpha) \mathcal{S}_{tgt} ^{t}
\end{equation}
where $\alpha$ is a smoothing coefficient ($\alpha$ = 0.999). Then, the teacher model can generate pseudo-labels to facilitate student model learning on the target domain.
To be mentioned, we notice that the direct usage of all pseudo-labels will lead to error accumulation in the teacher-student framework, especially in the initial cross-domain learning stage when the teacher model will commonly generate unreliable predictions due to domain shift. 
We thus utilize the spike-oriented uncertainty filter (same as Sec.~\ref{sec:cfkd}) to screen out unreliable pseudo labels, which aims to correct the mistakes that the teacher model made in cross-domain learning. 
As shown in Fig.~\ref{fig:method}, only the pseudo-labels ($\mathbf{D}_{pseudo}$ in Eq.~\ref{eq:TS}) with an uncertainty value below the threshold $\mathcal{E}^{var}$ are used to guide the student. This filtering mechanism greatly improves the upper bound of the teacher-student framework by ensuring the reliability of the teacher guidance.
Specifically, we use pseudo-labels as supervision and simply adopt the depth estimation SI loss in ~\cite{eigen2014depth}. The SI loss is shown below:
\begin{equation}
\scriptstyle
    \mathcal{L}_{\text{SCP}} = SI(\mathbf{D}_{pred}, \mathbf{D}_{pseudo}) = \frac{1}{W\times H} \sum_{i} d_{i}^{2}-\frac{\lambda}{(W\times H)^2}(\sum_{i}d_{i})^2
\label{eq:TS}
\end{equation}
where $d_{i}=\log\mathbf{D}_{pred}^{i}-\log\mathbf{D}_{pseudo}^{i}$, subscript $i$ indicates the pixel position, and $W$ and $H$ stand for width and height.

\textbf{Self-correcting alignment} \quad 
In order to further decrease the data distribution distance, we thus introduce the Self-correcting alignment mechanism to further avoid error accumulation in the process of domain adaptation. 
This mechanism can correct domain-invariant knowledge in the student model and transfer it to the teacher model progressively over the EMA process.
Specifically, as shown in Fig.~\ref{fig:IGKD} (a), this alignment is applied to the encoder outputs of the student model. 
We utilize global information to align the spike data distributions from two domains. Particularly, We extract class tokens as global vectors $\mathbf{f}_{src}, \mathbf{f}_{tgt}\in\mathbb{R}^{d}$ in two domains, which are further classified by two individual domain discriminators. 
The following equation represents the global-level spike alignment loss of the student model:
\begin{equation}
\scriptstyle
\mathcal{L}_{\text{SCA}} = \mathcal{L}_{adv}(\mathbf{f}_{src}, \mathbf{f}_{tgt}) = \log\mathcal{D}(\mathbf{f}_{src}) + \log(1-\mathcal{D}(\mathbf{f}_{tgt}))
\end{equation}
where $\mathcal{D}$ denotes the domain discriminator. 

\subsection{Loss Function}
\textbf{In cross-modality learning}, supervised training is applied on spike modality to ensure that the model $\mathcal{S}_{src}$ does not degenerate, and SI loss is also used ($\mathcal{L}_{\text{sup}}$). Specifically, the overall loss $\mathcal{L}_{\text{mod}}$ in this phase is shown as below:
\begin{equation}
\scriptstyle
    \mathcal{L}_{\text{mod}} = \mathcal{L}_{\text{sup}} + \mathcal{L}_{\text{unc}} + 0.1 * \mathcal{L}_{\text{CFKD}}
\label{eq:MOD}
\end{equation}

\textbf{In cross-domain learning}, we use pseudo-labels to supervise the spike student model, while also incorporating global-level alignment to jointly achieve self-correction for domain shift. The integrated loss $\mathcal{L}_{\text{dom}}$ consists of two parts:
\begin{equation}
\scriptstyle
    \mathcal{L}_{\text{dom}} =  \mathcal{L}_{\text{SCP}} + \mathcal{L}_{\text{unc}} + 0.1 *  \mathcal{L}_{\text{SCA}}
\label{eq:DOM}
\end{equation}

Although we can train cross-modality and cross-domain phases together, we adopt a separate training strategy, since it is of better performance compared with together training. During the evaluation stage, we only use target spike data as input, and infer on trained target spike model $\mathcal{S}_{tgt}$.

\section{Experiments}
In Sec~\ref{sec:4.1}, the setup of cross-modality cross-domain (BiCross) scenarios and implementation details are given.
In Sec~\ref{sec:4.2}, we evaluate the performance of our method in four challenging BiCross scenarios.
Comprehensive ablation studies are conducted in Sec~\ref{sec:4.3} to investigate the impact of each component.
Finally, we provide qualitative analysis to show the effectiveness of our method in Sec~\ref{sec:4.4}.

\begin{table}[t]
  \centering
  \caption{Results of different methods evaluated on Synthetic to Real scenario, from Virtual KITTI RGB to KITTI spike.} 
  \vspace{-0.25cm}
    \setlength{\tabcolsep}{1.0mm}{
    \begin{tabular}{c|c|c|cc}
    \hline
    Method & Train on & Pre-train & \multicolumn{1}{l}{Abs Rel ↓}  & \multicolumn{1}{l}{$\delta > {1.25}$ ↑} \\
    \hline\hline
    AdaDepth~\cite{kundu2018adadepth} & VKITTI-RGB & ImageNet & 0.303 & 0.506  \\
    T2Net~\cite{zheng2018t2net} & VKITTI-RGB & ImageNet & 0.252  & 0.689 \\
    SFA~\cite{wang2021exploring} & VKITTI-RGB & ImageNet & 0.298 & 0.541  \\
    \hline
    $Pre_{src}$ & VKITTI-spike & ImageNet & 0.285  & 0.663 \\
    \textbf{MonoCross} & VKITTI-spike & VKITTI-RGB & 0.217 & 0.704\\
    \textbf{BiCross} & VKITTI-spike &  MonoCross & \textbf{0.128} & \textbf{0.817} \\
    \hline
    \hline
    Supervised & KITTI-spike & ImageNet & 0.120 & 0.857 \\
    \hline
    \end{tabular}%
    }
  \label{tab:virtual}%
  \vspace{-0.1cm}
\end{table}%

\begin{table}[t]
  \centering
  \caption{Results of methods evaluated on Extreme Weather scenario, from Driving stereo clear RGB to foggy spike.} 
  \vspace{-0.25cm}
    \setlength{\tabcolsep}{1.0mm}{
    \begin{tabular}{c|c|c|cc}
    \hline
    Method & Train on & Pre-train & \multicolumn{1}{l}{Abs Rel ↓} & \multicolumn{1}{l}{$\delta > {1.25}$ ↑} \\
    \hline\hline
    AdaDepth~\cite{kundu2018adadepth} & Clear-RGB & ImageNet & 0.377 & 0.354  \\
    T2Net~\cite{zheng2018t2net} & Clear-RGB & ImageNet & 0.147 & 0.725  \\
    SFA~\cite{wang2021exploring} & Clear-RGB & ImageNet & 0.351 & 0.402 \\
    \hline
    $Pre_{src}$ & Clear-spike & ImageNet & 0.194  & 0.633  \\
    \textbf{MonoCross} & Clear-spike & Clear-RGB & 0.129 & 0.783\\
    \textbf{BiCross} & Clear-spike & MonoCross & \textbf{0.106} & \textbf{0.851} \\
    \hline
    Supervised & Foggy-spike & ImageNet & 0.120 & 0.864\\
    \hline
    \end{tabular}%
    }
  \label{tab:weather}%
  \vspace{-0.3cm}
\end{table}%

\begin{table}[t]
  \centering
  \caption{Results of different methods evaluated on Scene Changing scenario, from KITTI RGB to DrivingStereo spike.} 
  \vspace{-0.25cm}
    \setlength{\tabcolsep}{1.0mm}{
    \begin{tabular}{c|c|c|cc}
    \hline
    Method & Train on & Pre-train & \multicolumn{1}{l}{Abs Rel ↓}  & \multicolumn{1}{l}{$\delta > {1.25}$ ↑} \\
    \hline\hline
    $Pre_{src}$ & KITTI-spike & ImageNet & 0.322 & 0.160  \\
    \textbf{MonoCross} & KITTI-spike & KITTI-RGB & 0.293 & 0.183\\
    \textbf{BiCross} & KITTI-spike & MonoCross & \textbf{0.251}  & \textbf{0.309}\\
    \hline
    \end{tabular}%
    }
  \label{tab:scene}%
  \vspace{-0.1cm}
\end{table}%

\begin{table}[h]
  \centering
  \caption{Results of our proposed methods evaluated on Real Spike scenario, from NYUv2 RGB to Respike indoor spike data.} 
  \vspace{-0.25cm}
    \setlength{\tabcolsep}{1.2mm}{
    \begin{tabular}{c|c|c|cc}
    \hline
    Method & Train on & Pretrain & \multicolumn{1}{l}{Abs Rel ↓} & \multicolumn{1}{l}{$\delta > {1.25}$ ↑} \\
    \hline\hline
    $Pre_{src}$ & NYU-spike & ImageNet & 0.372 & 0.450 \\
    \textbf{MonoCross} & NYU-spike & NYU-RGB & 0.329 & 0.457 \\
    \textbf{BiCross} & NYU-spike &  MonoCross & \textbf{0.223} & \textbf{0.715} \\
    \hline
    \end{tabular}%
    }
  \label{tab:respike}%
  \vspace{-0.3cm}
\end{table}%

\subsection{Experimental Setup}
\label{sec:4.1}

\textbf{Data and label acquisition} \quad 
In our work, we generate spike datasets using RGB frames of three open access outdoor datasets, including  KITTI~\cite{geiger2012we}, Virtual KITTI~\cite{gaidon2016virtual}, and Driving Stereo (including different weathers)~\cite{yang2019drivingstereo}, and one indoor dataset NYUv2~\cite{silberman2012indoor}. Following previous work~\cite{hu2021optical, zhu2021neuspike}, we use a video frame insertion method~\cite{sim2021xvfi} and spike stream simulator to get spike data with 1280Hz, which reaches the real spike camera workflow. The dataset details are given in \href{https://github.com/Theia-4869/BiCross} {https://github.com/Theia-4869/BiCross}.

\textbf{BiCross scenarios} \quad 
In order to promote the development of neuromorphic spike cameras and evaluate the effectiveness of our method, we introduce four challenging BiCross scenarios:
\textbf{1) Synthetic to Real}, we set RGB Virtual KITTI as source data and realize the unsupervised depth estimation on target KITTI spike data. 
\textbf{2) Extreme Weather}, the RGB Driving Stereo in normal weather is considered as source data, and foggy spike Driving Stereo is considered as target data. 
\textbf{3) Scene Changing}, we design KITTI RGB as source data and transfer the knowledge to target spike Driving Stereo. Scene layouts are not static in real-world applications, especially in autonomous driving. 
\textbf{4) Real Spike}, we also adopt real spike data as the target domain, realizing BiCross from source NYU~\cite{silberman2012indoor} RGB to target Respike spike data~\cite{wang2022respike}.

\textbf{Implementation details} \quad 
BiCross framework is built based on DPT~\cite{ranftl2021vision}. We set ImageNet~\cite{deng2009imagenet} pre-trained ViT-Hybrid as transformer encoder in all experiments, whereas the decoder and prediction head are initialized randomly. The structures of depth and uncertainty estimation head are the same, which contain 3 convolutional layers. We set a learning rate of $1e-5$ for the backbone and $1e-4$ for the decoder. We adopt Adam optimizer~\cite{kingma2014adam}  $(\beta_1, \beta_2) = (0.9, 0.999)$ during cross-modality and cross-domain training for 30 and 10 epochs respectively. The batch size is set to 8 for all BiCross scenarios. For the input data, we first resize the longer side to 384 pixels and random crop a patch of 384 $\times$ 384. We only use random horizontal flips for RGB and spike data augmentation. The evaluation metrics are following previous depth estimation works~\cite{fu2018deep,ranftl2021vision}. All experiments are conducted on NVIDIA Tesla V100 GPUs.

\subsection{Effectiveness}
\label{sec:4.2}
In this section, we compare our method against the baselines \cite{ranftl2021vision, kundu2018adadepth, zheng2018t2net, wang2021exploring} and supervised method on four BiCross scenarios. In details of method setting, $Pre_{src}$ is directly trained on simulated source spike data, MonoCross is a part of BiCross which only adopts cross-modality training, BiCross is the entire process of our method, and Supervised is trained on target spike dataset with depth label. 
In addition, we further compare with previous unsupervised methods, including AdaDepth~\cite{kundu2018adadepth}, T2Net~\cite{zheng2018t2net}, and SFA~\cite{wang2021exploring} while altering their network to DPT. 
\textbf{Synthetic to Real.} As shown in Tab.~\ref{tab:virtual}, our method can outperform other unsupervised transferring methods (i.e., AdaDepth, T2Net, and SFA), since AdaDepth and SFA can hardly align the features under an enormous gap, and T2Net is difficult to translate input data style from RGB to spike modality. BiCross reduces 0.124 AbsRel and improves 12.8\% $\delta > {1.25}$ compared with the previous SOTA method. Meanwhile, MonoCross can exceed $Pre_{src}$ method in which CFKD reserves the advantages of two modalities and thus improves the performance on target spike data. 
Note that, BiCross also achieves competitive results compared with the supervised method. 
\textbf{Extreme Weather.} In order to verify the generalization of our method, we conduct experiments on the scenario of Extreme Weather changes. We set clear RGB as source data, and foggy spike streams as target data. As shown in Tab.~\ref{tab:weather}, BiCross achieves higher accuracy than other previous unsupervised methods and $Pre_{src}$, showing consistent performance as Synthetic to Real scenario. To be mentioned, due to the small quantity of severe weather target data, BiCross can outperform the supervised method by 0.014 AbsRel in foggy target data, showing the great potential of our method. 
Besides, we conduct another extreme weather-changing experiment in \href{https://github.com/Theia-4869/BiCross} {code web}.
\textbf{Scene Changing} and \textbf{Real Spike.} BiCross still outperforms other methods by a considerable margin, as shown in Tab.~\ref{tab:scene} and Tab.~\ref{tab:respike}. Particularly, BiCross decreases 0.071 and 0.149 AbsRel compared with $Pre_{src}$ in these two BiCross scenarios. The results demonstrate that our framework can realize stable depth estimation in unlabeled target spike data, no matter if the spike data is simulated or captured from the real world.

\subsection{Ablation Study}
\label{sec:4.3}
We evaluate the contribution of each component on the Synthetic to Real scenario. 
We divide CFKD into coarse (CKD) and fine (FKD) distillation, and SCTS into self-correcting pseudo-label (SCP) and alignment (SCA). 

\textbf{Effectiveness of each component} As presented in Tab.~\ref{tab:abl}, the first row showcases the performance of the source RGB pre-trained model. 
In the cross-modality phase, $Ex_{1}$ verifies the effectiveness of CFKD, and $Ex_{2}$ further proves the effectiveness of the filtering mechanism in distillation, which outperforms RGB by 0.077 AbsRel.
And the filtering mechanism gains an extra 0.026 AbsRel improvement.
This means the cross-modality phase can provide more semantic information to the student model and improve the generalization ability of the model on the target spike. 
In the cross-domain phase, without cross-modality knowledge distillation, we directly evaluate SCTS on $Ex_{3}$ and $Ex_{4}$, which achieved promising results on the target spike. Compared with RGB, $Ex_{3}$ decreases the AbsRel from 0.294 to 0.261, and $Ex_{4}$ decreases the AbsRel to 0.205. The results demonstrate that SCTS can further address the domain shift, and the self-correcting scheme is crucial in pixel-wise cross-domain learning.
From $Ex_{5}$ to $Ex_{8}$, based on CFKD, we progressively add SCTS and the uncertainty filter throughout the training. $Ex_{8}$ achieves the best depth estimation accuracy, which verifies the effectiveness of each component in BiCross. 

\begin{table}[t]
  \centering
  \caption{Ablation studies on the Virtual KITTI RGB to KITTI spike. It shows the effectiveness of CFKD, SCTS, and spike-oriented uncertainty Filter. \emptyCircle\ represents exclusion, \fullCircle\ represents inclusion, and \halfCircle\ represents inclusion with an uncertainty filter. RGB stands for source RGB pre-trained. Modality, Domain, and Both mean the cross-modality, cross-domain, and entire BiCross phase, respectively.}
    \vspace{-0.25cm}
    \setlength{\tabcolsep}{1.2mm}{
    \begin{tabular}{c|cc|cc}
    \hline
    Phase & CFKD & SCTS & \multicolumn{1}{l}{Abs Rel ↓} & \multicolumn{1}{l}{$\delta > {1.25}$ ↑} \\
    \hline
    \hline
    RGB & \emptyCircle & \emptyCircle & 0.294 & 0.658 \\
    \hline
    Modality ($Ex_{1}$) & \fullCircle & \emptyCircle & 0.243 & 0.702 \\
    Modality ($Ex_{2}$) & \halfCircle & \emptyCircle & 0.217 & 0.704 \\
    Domain ($Ex_{3}$) & \emptyCircle & \fullCircle & 0.261 & 0.685  \\
    Domain ($Ex_{4}$) & \emptyCircle & \halfCircle & 0.205 & 0.740 \\
    Both ($Ex_{5}$) & \fullCircle & \fullCircle & 0.216 & 0.736 \\
    Both ($Ex_{6}$) & \halfCircle & \fullCircle & 0.197 & 0.741 \\
    Both ($Ex_{7}$) & \fullCircle & \halfCircle & 0.129 & 0.810 \\
    Both ($Ex_{8}$) & \halfCircle & \halfCircle & 0.128 & 0.817 \\
    \hline  
    \end{tabular}}
  \label{tab:abl}
\end{table}

\begin{table}[t]
  \centering
  \caption{Ablation studies on the effectiveness of subdivided modules in CFKD. The settings and notations are in accordance with Tab.~\ref{tab:abl}. All experiments are conducted in the cross-modality phase and source RGB pre-trained weights are used.}
    \vspace{-0.25cm}
    \setlength{\tabcolsep}{1.2mm}{
    \begin{tabular}{c|cc|cc}
    \hline
    Phase & CKD & FKD & \multicolumn{1}{l}{Abs Rel ↓} & \multicolumn{1}{l}{$\delta > {1.25}$ ↑} \\
    \hline
    \hline
    RGB & \emptyCircle & \emptyCircle & 0.294 & 0.658 \\
    \hline
    Modality ($Ex_{9}$) & \halfCircle & \emptyCircle & 0.239 & 0.686 \\
    Modality ($Ex_{10}$) & \emptyCircle & \halfCircle & 0.246 & 0.683 \\
    Modality ($Ex_{2}$) & \halfCircle & \halfCircle & 0.217 & 0.704 \\
    \hline  
    \end{tabular}}
  \label{tab:abl_cfkd}
  \vspace{-0.3cm}
\end{table}

\begin{table}[t]
  \centering
  \caption{Ablation studies on the effectiveness of subdivided modules in SCTS. The settings and notations are in accordance with Tab.~\ref{tab:abl}. All experiments are conducted in the cross-domain phase and the complete CFKD was included.}
    \vspace{-0.25cm}
    \setlength{\tabcolsep}{1.2mm}{
    \begin{tabular}{c|cc|cc}
    \hline
    Phase & SCP & SCA & \multicolumn{1}{l}{Abs Rel ↓} & \multicolumn{1}{l}{$\delta > {1.25}$ ↑} \\
    \hline
    \hline
    Modality ($Ex_{2}$) & \emptyCircle & \emptyCircle & 0.217 & 0.704 \\
    \hline
    Both ($Ex_{11}$) & \halfCircle & \emptyCircle & 0.170 & 0.757 \\
    Both ($Ex_{12}$) & \emptyCircle & \fullCircle & 0.159 & 0.796 \\
    Both ($Ex_{8}$) & \halfCircle & \fullCircle & 0.128 & 0.817 \\
    \hline  
    \end{tabular}}
  \label{tab:abl_scts}
\end{table}

\textbf{Effectiveness of sub-component} 
In this part, we demonstrate the effectiveness of each sub-component in each phase.
As shown in Tab.~\ref{tab:abl_cfkd}, $Ex_{9}$ and $Ex_{10}$ verify the effectiveness of the CKD and FKD module respectively. Their combination $Ex_{2}$, compared to each of the two modules separately, reduces the AbsRel to 0.217, which further demonstrates that they jointly facilitate in cross-modality learning phase.
Furthermore, the sub-components of SCTS are shown in Tab.~\ref{tab:abl_scts}. The self-correcting pseudo-label ($Ex_{11}$) and alignment ($Ex_{12}$) can respectively improve the adaptability of the model, and the overall SCTS mechanism ($Ex_{8}$) can further reduce the AbsRel to 0.128 under unsupervised setting.

\subsection{Qualitative Analysis}
\label{sec:4.4}
\begin{figure}[t]
\includegraphics[width=0.47\textwidth]{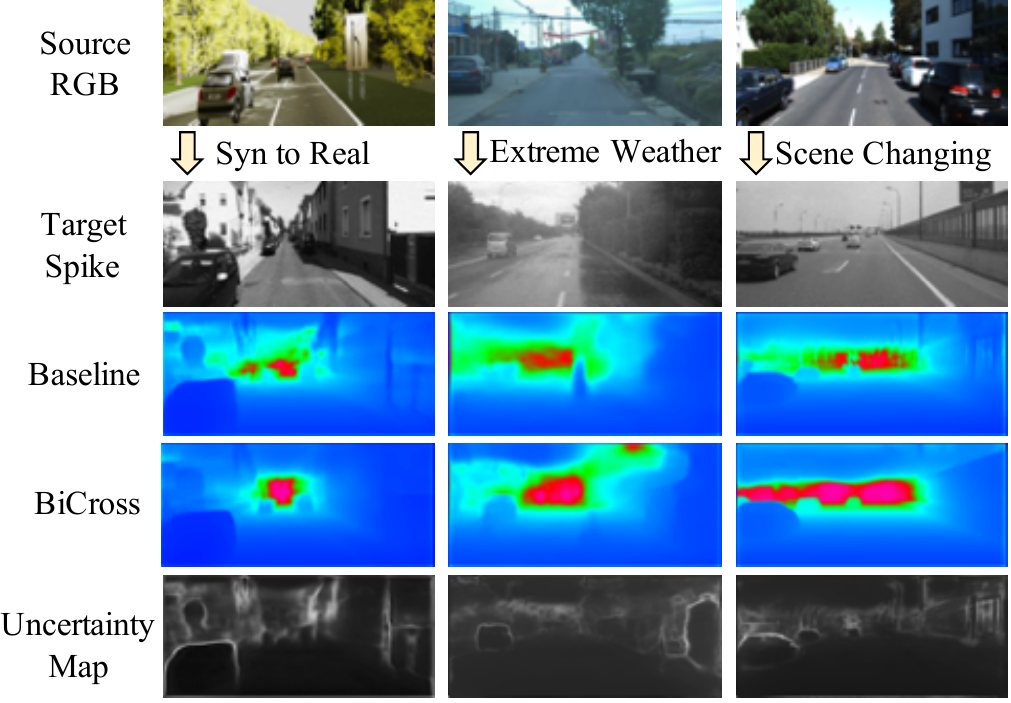}
\centering
\vspace{-0.1cm}
\caption{Baseline represents the output of $Pre_{src}$. Uncertainty Map is predicted by our method in unsupervised cross-domain phase.}
\label{fig:Vis}
\vspace{-0.3cm}
\end{figure}

We show the qualitative comparison of outputs in Fig.~\ref{fig:Vis}. As can be seen, our method achieves better depth maps compared with the baseline in three scenarios, which demonstrates that our method can effectively realize unsupervised spike depth estimation on spike data. In the last row, the uncertainty map for the pixel-level filter also achieves satisfying results, showing a higher uncertainty value on the edge of objects.

\section{Conclusion}
We are the first to explore the unsupervised task in spike modality, and propose a BiCross framework to leverage the annotation and absorb sufficient spatial information from open-source RGB datasets. For the cross-modality phase, Coarse-to-Fine Knowledge Distillation is designed to realize comprehensive cross-modality knowledge transfer and reserve the unique strength of both modalities. For the cross-domain phase, we introduce a Self-Correcting Teacher-Student scheme to ease the domain shift and avoid error accumulation. We provide four large-scale spike
depth estimation datasets for continuous research in the spike community.

\section{Acknowledgement}
Shanghang Zhang is supported by the National
Key Research and Development Project of China
(No.2022ZD0117801).

{
\bibliographystyle{IEEEtran}
\bibliography{IEEEabrv,reference}

\begin{thebibliography}{10}
\providecommand{\url}[1]{#1}
\csname url@samestyle\endcsname
\providecommand{\newblock}{\relax}
\providecommand{\bibinfo}[2]{#2}
\providecommand{\BIBentrySTDinterwordspacing}{\spaceskip=0pt\relax}
\providecommand{\BIBentryALTinterwordstretchfactor}{4}
\providecommand{\BIBentryALTinterwordspacing}{\spaceskip=\fontdimen2\font plus
\BIBentryALTinterwordstretchfactor\fontdimen3\font minus \fontdimen4\font\relax}
\providecommand{\BIBforeignlanguage}[2]{{%
\expandafter\ifx\csname l@#1\endcsname\relax
\typeout{** WARNING: IEEEtran.bst: No hyphenation pattern has been}%
\typeout{** loaded for the language `#1'. Using the pattern for}%
\typeout{** the default language instead.}%
\else
\language=\csname l@#1\endcsname
\fi
#2}}
\providecommand{\BIBdecl}{\relax}
\BIBdecl

\bibitem{dong2019efficient}
S.~Dong, L.~Zhu, D.~Xu, Y.~Tian, and T.~Huang, ``An efficient coding method for spike camera using inter-spike intervals,'' \emph{arXiv preprint arXiv:1912.09669}, 2019.

\bibitem{zhu2019retina}
L.~Zhu, S.~Dong, T.~Huang, and Y.~Tian, ``A retina-inspired sampling method for visual texture reconstruction,'' in \emph{2019 IEEE International Conference on Multimedia and Expo (ICME)}.\hskip 1em plus 0.5em minus 0.4em\relax IEEE, 2019, pp. 1432--1437.

\bibitem{manhardt2019roi}
F.~Manhardt, W.~Kehl, and A.~Gaidon, ``Roi-10d: Monocular lifting of 2d detection to 6d pose and metric shape,'' in \emph{Proceedings of the IEEE/CVF Conference on Computer Vision and Pattern Recognition}, 2019, pp. 2069--2078.

\bibitem{wu20196d}
D.~Wu, Z.~Zhuang, C.~Xiang, W.~Zou, and X.~Li, ``6d-vnet: End-to-end 6-dof vehicle pose estimation from monocular rgb images,'' in \emph{Proceedings of the IEEE/CVF Conference on Computer Vision and Pattern Recognition Workshops}, 2019, pp. 0--0.

\bibitem{chi2023bev}
X.~Chi, J.~Liu, M.~Lu, R.~Zhang, Z.~Wang, Y.~Guo, and S.~Zhang, ``Bev-san: Accurate bev 3d object detection via slice attention networks,'' in \emph{Proceedings of the IEEE/CVF Conference on Computer Vision and Pattern Recognition}, 2023, pp. 17\,461--17\,470.

\bibitem{li2023bev}
J.~Li, M.~Lu, J.~Liu, Y.~Guo, Y.~Du, L.~Du, and S.~Zhang, ``Bev-lgkd: A unified lidar-guided knowledge distillation framework for multi-view bev 3d object detection,'' \emph{IEEE Transactions on Intelligent Vehicles}, 2023.

\bibitem{yang2023lidar}
S.~Yang, J.~Liu, R.~Zhang, M.~Pan, Z.~Guo, X.~Li, Z.~Chen, P.~Gao, Y.~Guo, and S.~Zhang, ``Lidar-llm: Exploring the potential of large language models for 3d lidar understanding,'' \emph{arXiv preprint arXiv:2312.14074}, 2023.

\bibitem{tremblay2018deep}
J.~Tremblay, T.~To, B.~Sundaralingam, Y.~Xiang, D.~Fox, and S.~Birchfield, ``Deep object pose estimation for semantic robotic grasping of household objects,'' \emph{arXiv preprint arXiv:1809.10790}, 2018.

\bibitem{li2023manipllm}
X.~Li, M.~Zhang, Y.~Geng, H.~Geng, Y.~Long, Y.~Shen, R.~Zhang, J.~Liu, and H.~Dong, ``Manipllm: Embodied multimodal large language model for object-centric robotic manipulation,'' \emph{arXiv preprint arXiv:2312.16217}, 2023.

\bibitem{li2023imagemanip}
X.~Li, Y.~Wang, Y.~Shen, P.~Iaroslav, H.~Lu, Q.~Wang, B.~An, J.~Liu, and H.~Dong, ``Imagemanip: Image-based robotic manipulation with affordance-guided next view selection,'' \emph{arXiv preprint arXiv:2310.09069}, 2023.

\bibitem{hu2021optical}
L.~Hu, R.~Zhao, Z.~Ding, L.~Ma, B.~Shi, R.~Xiong, and T.~Huang, ``Optical flow estimation for spiking camera,'' \emph{arXiv preprint arXiv:2110.03916}, 2021.

\bibitem{hu2014joint}
Z.~Hu, L.~Xu, and M.-H. Yang, ``Joint depth estimation and camera shake removal from single blurry image,'' in \emph{Proceedings of the IEEE Conference on Computer Vision and Pattern Recognition}, 2014, pp. 2893--2900.

\bibitem{dong2021spike}
S.~Dong, T.~Huang, and Y.~Tian, ``Spike camera and its coding methods,'' \emph{arXiv preprint arXiv:2104.04669}, 2021.

\bibitem{li2022retinomorphic}
J.~Li, X.~Wang, L.~Zhu, J.~Li, T.~Huang, and Y.~Tian, ``Retinomorphic object detection in asynchronous visual streams,'' 2022.

\bibitem{zhang2022spike}
J.~Zhang, L.~Tang, Z.~Yu, J.~Lu, and T.~Huang, ``Spike transformer: Monocular depth estimation for spiking camera,'' in \emph{Computer Vision--ECCV 2022: 17th European Conference, Tel Aviv, Israel, October 23--27, 2022, Proceedings, Part VII}.\hskip 1em plus 0.5em minus 0.4em\relax Springer, 2022, pp. 34--52.

\bibitem{godard2017unsupervised}
C.~Godard, O.~Mac~Aodha, and G.~J. Brostow, ``Unsupervised monocular depth estimation with left-right consistency,'' in \emph{Proceedings of the IEEE conference on computer vision and pattern recognition}, 2017, pp. 270--279.

\bibitem{lopez2020desc}
A.~Lopez-Rodriguez and K.~Mikolajczyk, ``Desc: Domain adaptation for depth estimation via semantic consistency,'' \emph{arXiv preprint arXiv:2009.01579}, 2020.

\bibitem{gaidon2016virtual}
A.~Gaidon, Q.~Wang, Y.~Cabon, and E.~Vig, ``Virtual worlds as proxy for multi-object tracking analysis,'' in \emph{Proceedings of the IEEE conference on computer vision and pattern recognition}, 2016, pp. 4340--4349.

\bibitem{yang2019drivingstereo}
G.~Yang, X.~Song, C.~Huang, Z.~Deng, J.~Shi, and B.~Zhou, ``Drivingstereo: A large-scale dataset for stereo matching in autonomous driving scenarios,'' in \emph{Proceedings of the IEEE/CVF Conference on Computer Vision and Pattern Recognition}, 2019, pp. 899--908.

\bibitem{geiger2012we}
A.~Geiger, P.~Lenz, and R.~Urtasun, ``Are we ready for autonomous driving? the kitti vision benchmark suite,'' in \emph{2012 IEEE conference on computer vision and pattern recognition}.\hskip 1em plus 0.5em minus 0.4em\relax IEEE, 2012, pp. 3354--3361.

\bibitem{silberman2012indoor}
N.~Silberman, D.~Hoiem, P.~Kohli, and R.~Fergus, ``Indoor segmentation and support inference from rgbd images,'' in \emph{European conference on computer vision}.\hskip 1em plus 0.5em minus 0.4em\relax Springer, 2012, pp. 746--760.

\bibitem{wang2022respike}
Y.~Wang, J.~Li, L.~Zhu, X.~Xiang, T.~Huang, and Y.~Tian, ``Learning stereo depth estimation with bio-inspired spike cameras,'' in \emph{2022 IEEE International Conference on Multimedia and Expo (ICME)}, 2022, pp. 1--6.

\bibitem{eigen2014depth}
D.~Eigen, C.~Puhrsch, and R.~Fergus, ``Depth map prediction from a single image using a multi-scale deep network,'' \emph{Advances in neural information processing systems}, vol.~27, 2014.

\bibitem{Silberman:ECCV12}
P.~K. Nathan~Silberman, Derek~Hoiem and R.~Fergus, ``Indoor segmentation and support inference from rgbd images,'' in \emph{ECCV}, 2012.

\bibitem{scharstein2014high}
D.~Scharstein, H.~Hirschm{\"u}ller, Y.~Kitajima, G.~Krathwohl, N.~Ne{\v{s}}i{\'c}, X.~Wang, and P.~Westling, ``High-resolution stereo datasets with subpixel-accurate ground truth,'' in \emph{German conference on pattern recognition}.\hskip 1em plus 0.5em minus 0.4em\relax Springer, 2014, pp. 31--42.

\bibitem{Xu_2018_CVPR}
D.~Xu, W.~Wang, H.~Tang, H.~Liu, N.~Sebe, and E.~Ricci, ``Structured attention guided convolutional neural fields for monocular depth estimation,'' in \emph{Proceedings of the IEEE Conference on Computer Vision and Pattern Recognition (CVPR)}, June 2018.

\bibitem{Ramamonjisoa_2020_CVPR}
M.~Ramamonjisoa, Y.~Du, and V.~Lepetit, ``Predicting sharp and accurate occlusion boundaries in monocular depth estimation using displacement fields,'' in \emph{Proceedings of the IEEE/CVF Conference on Computer Vision and Pattern Recognition (CVPR)}, June 2020.

\bibitem{lee2019monocular}
J.-H. Lee and C.-S. Kim, ``Monocular depth estimation using relative depth maps,'' in \emph{Proceedings of the IEEE/CVF Conference on Computer Vision and Pattern Recognition}, 2019, pp. 9729--9738.

\bibitem{Ramamonjisoa_2019_ICCV}
M.~Ramamonjisoa and V.~Lepetit, ``Sharpnet: Fast and accurate recovery of occluding contours in monocular depth estimation,'' in \emph{Proceedings of the IEEE/CVF International Conference on Computer Vision (ICCV) Workshops}, Oct 2019.

\bibitem{fu2018deep}
H.~Fu, M.~Gong, C.~Wang, K.~Batmanghelich, and D.~Tao, ``Deep ordinal regression network for monocular depth estimation,'' in \emph{Proceedings of the IEEE conference on computer vision and pattern recognition}, 2018, pp. 2002--2011.

\bibitem{garg2016unsupervised}
R.~Garg, V.~K. Bg, G.~Carneiro, and I.~Reid, ``Unsupervised cnn for single view depth estimation: Geometry to the rescue,'' in \emph{European conference on computer vision}.\hskip 1em plus 0.5em minus 0.4em\relax Springer, 2016, pp. 740--756.

\bibitem{zhou2017unsupervised}
T.~Zhou, M.~Brown, N.~Snavely, and D.~G. Lowe, ``Unsupervised learning of depth and ego-motion from video,'' in \emph{Proceedings of the IEEE conference on computer vision and pattern recognition}, 2017, pp. 1851--1858.

\bibitem{godard2019digging}
C.~Godard, O.~Mac~Aodha, M.~Firman, and G.~J. Brostow, ``Digging into self-supervised monocular depth estimation,'' in \emph{Proceedings of the IEEE/CVF International Conference on Computer Vision}, 2019, pp. 3828--3838.

\bibitem{pilzer2019refine}
A.~Pilzer, S.~Lathuiliere, N.~Sebe, and E.~Ricci, ``Refine and distill: Exploiting cycle-inconsistency and knowledge distillation for unsupervised monocular depth estimation,'' in \emph{Proceedings of the IEEE/CVF Conference on Computer Vision and Pattern Recognition}, 2019, pp. 9768--9777.

\bibitem{chen2021revealing}
Z.~Chen, X.~Ye, W.~Yang, Z.~Xu, X.~Tan, Z.~Zou, E.~Ding, X.~Zhang, and L.~Huang, ``Revealing the reciprocal relations between self-supervised stereo and monocular depth estimation,'' in \emph{Proceedings of the IEEE/CVF International Conference on Computer Vision}, 2021, pp. 15\,529--15\,538.

\bibitem{piao2021learning}
Y.~Piao, X.~Ji, M.~Zhang, and Y.~Zhang, ``Learning multi-modal information for robust light field depth estimation,'' \emph{arXiv preprint arXiv:2104.05971}, 2021.

\bibitem{verdie2022cromo}
Y.~Verdi{\'e}, J.~Song, B.~Mas, B.~Busam, A.~Leonardis, and S.~McDonagh, ``Cromo: Cross-modal learning for monocular depth estimation,'' \emph{arXiv preprint arXiv:2203.12485}, 2022.

\bibitem{wang2021evdistill}
L.~Wang, Y.~Chae, S.-H. Yoon, T.-K. Kim, and K.-J. Yoon, ``Evdistill: Asynchronous events to end-task learning via bidirectional reconstruction-guided cross-modal knowledge distillation,'' in \emph{Proceedings of the IEEE/CVF Conference on Computer Vision and Pattern Recognition}, 2021, pp. 608--619.

\bibitem{kundu2018adadepth}
J.~N. Kundu, P.~K. Uppala, A.~Pahuja, and R.~V. Babu, ``Adadepth: Unsupervised content congruent adaptation for depth estimation,'' in \emph{Proceedings of the IEEE conference on computer vision and pattern recognition}, 2018, pp. 2656--2665.

\bibitem{zheng2018t2net}
C.~Zheng, T.-J. Cham, and J.~Cai, ``T2net: Synthetic-to-realistic translation for solving single-image depth estimation tasks,'' in \emph{Proceedings of the European conference on computer vision (ECCV)}, 2018, pp. 767--783.

\bibitem{zhao2019geometry}
S.~Zhao, H.~Fu, M.~Gong, and D.~Tao, ``Geometry-aware symmetric domain adaptation for monocular depth estimation,'' in \emph{Proceedings of the IEEE/CVF Conference on Computer Vision and Pattern Recognition}, 2019, pp. 9788--9798.

\bibitem{zhu2020retina}
L.~Zhu, S.~Dong, J.~Li, T.~Huang, and Y.~Tian, ``Retina-like visual image reconstruction via spiking neural model,'' in \emph{Proceedings of the IEEE/CVF Conference on Computer Vision and Pattern Recognition}, 2020, pp. 1438--1446.

\bibitem{9181055}
J.~Zhao, R.~Xiong, and T.~Huang, ``High-speed motion scene reconstruction for spike camera via motion aligned filtering,'' in \emph{2020 IEEE International Symposium on Circuits and Systems (ISCAS)}, 2020, pp. 1--5.

\bibitem{zheng2021high}
Y.~Zheng, L.~Zheng, Z.~Yu, B.~Shi, Y.~Tian, and T.~Huang, ``High-speed image reconstruction through short-term plasticity for spiking cameras,'' in \emph{Proceedings of the IEEE/CVF Conference on Computer Vision and Pattern Recognition}, 2021, pp. 6358--6367.

\bibitem{zhu2021neuspike}
L.~Zhu, J.~Li, X.~Wang, T.~Huang, and Y.~Tian, ``Neuspike-net: High speed video reconstruction via bio-inspired neuromorphic cameras,'' in \emph{Proceedings of the IEEE/CVF International Conference on Computer Vision}, 2021, pp. 2400--2409.

\bibitem{zhao2021super}
J.~Zhao, J.~Xie, R.~Xiong, J.~Zhang, Z.~Yu, and T.~Huang, ``Super resolve dynamic scene from continuous spike streams,'' in \emph{Proceedings of the IEEE/CVF International Conference on Computer Vision}, 2021, pp. 2533--2542.

\bibitem{zhao2021spk2imgnet}
J.~Zhao, R.~Xiong, H.~Liu, J.~Zhang, and T.~Huang, ``Spk2imgnet: Learning to reconstruct dynamic scene from continuous spike stream,'' in \emph{Proceedings of the IEEE/CVF Conference on Computer Vision and Pattern Recognition}, 2021, pp. 11\,996--12\,005.

\bibitem{zhu2022ultra}
L.~Zhu, S.~Dong, J.~Li, T.~Huang, and Y.~Tian, ``Ultra-high temporal resolution visual reconstruction from a fovea-like spike camera via spiking neuron model,'' \emph{IEEE Transactions on Pattern Analysis and Machine Intelligence}, 2022.

\bibitem{hu2021scflow}
L.~Hu, R.~Zhao, Z.~Ding, R.~Xiong, L.~Ma, and T.~Huang, ``Scflow: Optical flow estimation for spiking camera,'' \emph{arXiv preprint arXiv:2110.03916}, 2021.

\bibitem{delbruck2010activity}
T.~Delbr{\"u}ck, B.~Linares-Barranco, E.~Culurciello, and C.~Posch, ``Activity-driven, event-based vision sensors,'' in \emph{Proceedings of 2010 IEEE international symposium on circuits and systems}.\hskip 1em plus 0.5em minus 0.4em\relax IEEE, 2010, pp. 2426--2429.

\bibitem{lichtsteiner2008128}
P.~Lichtsteiner, C.~Posch, and T.~Delbruck, ``A 128 x 128 120 db 15x1e-6 s latency asynchronous temporal contrast vision sensor,'' \emph{IEEE journal of solid-state circuits}, vol.~43, no.~2, pp. 566--576, 2008.

\bibitem{Zhao_2021_CVPR}
J.~Zhao, R.~Xiong, H.~Liu, J.~Zhang, and T.~Huang, ``Spk2imgnet: Learning to reconstruct dynamic scene from continuous spike stream,'' in \emph{Proceedings of the IEEE/CVF Conference on Computer Vision and Pattern Recognition (CVPR)}, June 2021, pp. 11\,996--12\,005.

\bibitem{wang2022learning}
Y.~Wang, J.~Li, L.~Zhu, X.~Xiang, T.~Huang, and Y.~Tian, ``Learning stereo depth estimation with bio-inspired spike cameras,'' in \emph{2022 IEEE International Conference on Multimedia and Expo (ICME)}.\hskip 1em plus 0.5em minus 0.4em\relax IEEE, 2022, pp. 1--6.

\bibitem{ranftl2021vision}
R.~Ranftl, A.~Bochkovskiy, and V.~Koltun, ``Vision transformers for dense prediction,'' in \emph{Proceedings of the IEEE/CVF International Conference on Computer Vision}, 2021, pp. 12\,179--12\,188.

\bibitem{he2016deep}
K.~He, X.~Zhang, S.~Ren, and J.~Sun, ``Deep residual learning for image recognition,'' in \emph{Proceedings of the IEEE conference on computer vision and pattern recognition}, 2016, pp. 770--778.

\bibitem{yu2022cross}
J.~Yu, J.~Liu, X.~Wei, H.~Zhou, Y.~Nakata, D.~Gudovskiy, T.~Okuno, J.~Li, K.~Keutzer, and S.~Zhang, ``Cross-domain object detection with mean-teacher transformer,'' \emph{arXiv preprint arXiv:2205.01643}, 2022.

\bibitem{pham2021meta}
H.~Pham, Z.~Dai, Q.~Xie, and Q.~V. Le, ``Meta pseudo labels,'' in \emph{Proceedings of the IEEE/CVF conference on computer vision and pattern recognition}, 2021, pp. 11\,557--11\,568.

\bibitem{tarvainen2017mean}
\BIBentryALTinterwordspacing
A.~Tarvainen and H.~Valpola, ``Mean teachers are better role models: Weight-averaged consistency targets improve semi-supervised deep learning results,'' in \emph{Advances in Neural Information Processing Systems}, I.~Guyon, U.~V. Luxburg, S.~Bengio, H.~Wallach, R.~Fergus, S.~Vishwanathan, and R.~Garnett, Eds., vol.~30.\hskip 1em plus 0.5em minus 0.4em\relax Curran Associates, Inc., 2017. [Online]. Available: \url{https://proceedings.neurips.cc/paper/2017/file/68053af2923e00204c3ca7c6a3150cf7-Paper.pdf}
\BIBentrySTDinterwordspacing

\bibitem{wang2021exploring}
W.~Wang, Y.~Cao, J.~Zhang, F.~He, Z.-J. Zha, Y.~Wen, and D.~Tao, ``Exploring sequence feature alignment for domain adaptive detection transformers,'' in \emph{Proceedings of the 29th ACM International Conference on Multimedia}, 2021, pp. 1730--1738.

\bibitem{sim2021xvfi}
H.~Sim, J.~Oh, and M.~Kim, ``Xvfi: Extreme video frame interpolation,'' in \emph{Proceedings of the IEEE/CVF International Conference on Computer Vision}, 2021, pp. 14\,489--14\,498.

\bibitem{deng2009imagenet}
J.~Deng, W.~Dong, R.~Socher, L.-J. Li, K.~Li, and L.~Fei-Fei, ``Imagenet: A large-scale hierarchical image database,'' in \emph{2009 IEEE conference on computer vision and pattern recognition}.\hskip 1em plus 0.5em minus 0.4em\relax Ieee, 2009, pp. 248--255.

\bibitem{kingma2014adam}
D.~P. Kingma and J.~Ba, ``Adam: A method for stochastic optimization,'' \emph{arXiv preprint arXiv:1412.6980}, 2014.

\end{thebibliography}
}

\end{document}